\UseRawInputEncoding

\documentclass[journal]{IEEEtran}
\ifCLASSINFOpdf
\else
\fi

\usepackage[pdftex]{graphicx}
\usepackage{subfigure}
\usepackage{footmisc}
\usepackage{url}
\usepackage{amssymb}
\usepackage{diagbox}
\usepackage{multirow}
\usepackage{array}
\usepackage{marvosym}
\usepackage{textcomp}
\usepackage{makecell}
\usepackage{booktabs}
\usepackage{epstopdf}
\usepackage{cite}
\usepackage{amsmath}
\usepackage{changes}

\begin{document}
%
\title{CrackCLF: Automatic Pavement Crack Detection based on Closed-Loop Feedback }
%
%

\author{Chong Li, Zhun Fan*, \emph{Senior Member, IEEE},  Ying Chen, Huibiao Lin, Laura Moretti, Giuseppe Loprencipe*, Weihua Sheng, \emph{Senior Member, IEEE}, Kelvin C. P. Wang

\thanks{This work was supported by the Science and Technology Planning Project of Guangdong Province of China under grant 180917144960530, by the Project of Educational Commission of Guangdong Province of China under grant 2017KZDXM032, by the Project of Robot Automatic Design Platform combining Multi-Objective Evolutionary Computation and Deep Neural Network under grant 2019A050519008, and by the State Key Lab of Digital Manufacturing Equipment and Technology under grant DMETKF2019020.}

\thanks{Chong Li, Zhun Fan, Ying Chen and Huibiao Lin are with the Key Lab of Digital Signal and Image Processing of Guangdong Province, College of Engineering, Shantou University, Shan¡¯tou 515063, China (email: 15cli@stu.edu.cn, zfan@stu.edu.cn, 1316957264@qq.com, 13hblin@stu.edu.cn).}

\thanks{Laura Moretti and Giuseppe Loprencipe are with the Department of Civil, Construction and Environmental Engineering, Sapienza University of Rome, 00184 Rome, Italy (email: laura.moretti,giuseppe.loprencipe@uniroma1.it).}

\thanks{Weihua Sheng is with the School of Electrical and Computer Engineering, Oklahoma State University, Stillwater, OK 74078 USA (email: weihua.sheng@okstate.edu).}

\thanks{Kelvin C. P. Wang is with the School of Civil and Environmental Engineering, Oklahoma State University, Stillwater, OK 74078 USA (email: kcpwang@gmail.com).}

\thanks{Corresponding authors: Zhun Fan, Giuseppe Loprencipe.}

\thanks{Digital Object Identifier 10.1109/TITS.2023.3332995.}
}

\markboth{IEEE TRANSACTIONS ON INTELLIGENT TRANSPORTATION SYSTEMS}%
{CrackCLF: Automatic Pavement Crack Detection based on Closed-Loop Feedback}
\maketitle

\begin{abstract}
Automatic pavement crack detection is an important task to ensure the functional performances of pavements during their service life. Inspired by deep learning (DL), the encoder-decoder framework is a powerful tool for crack detection. However, these models are usually open-loop (OL) systems that tend to treat thin cracks as the background. Meanwhile, these models can not automatically correct errors in the prediction, nor can it adapt to the changes of the environment to automatically extract and detect thin cracks. To tackle this problem, we embed closed-loop feedback (CLF) into the neural network so that the model could learn to correct errors on its own, based on generative adversarial networks (GAN). The resulting model is called CrackCLF and includes the front and back ends, i.e.  segmentation and adversarial network. The front end with U-shape framework is employed to generate crack maps, and the back end with a multi-scale loss function is used to correct higher-order inconsistencies between labels and crack maps (generated by the front end) to address open-loop system issues.
Empirical results show that the proposed CrackCLF outperforms others methods on three public datasets. Moreover, the proposed CLF can be defined as a plug and play module, which can be embedded into different neural network models to improve their performances.

\end{abstract}


%
\IEEEpeerreviewmaketitle

\section{Introduction}
%
%
%
%

\IEEEPARstart{P}avement crack is a common road distress, mainly caused by temperature, materials¡¯ aging, fatigue, and traffic loads \cite{amhaz2016automatic}. They often start from the bottom of the upper layers and propagate to surfaces \cite{2008Pavement}. These discontinuities affect the structures and functional performances of the pavement and shorten its service life, causing discomfort for road users and costs for road managers \cite{zaniewski1999pavement}. Moreover, there are potential threats to road safety with social costs\cite{zou2018deepcrack}: cracks may be penetrated by rain, forming potholes, which can cause accidents \cite{chan2010investigating}. Crack detection is a necessary step for road management to repair them, prevent their diffusion, and maintain structural and functional conditions of the pavement \cite{cubero2017efficient}.

In the past few decades, many structural health monitoring research were reported \cite{Park2010A,O2014Regionally}. The development and application of intelligent robotic systems for defect detection in civil infrastructure is advancing rapidly.
Oh et al. proposed a bridge detection system, including a specially designed car, a robot mechanism, and a machine vision system \cite{Oh2009Bridge}. Lim proposed a crack detection system that uses a camera to collect images. Laplacian of Gaussian was employed to inspect crack information, while camera calibration and robot localization were applied to obtain a global crack map \cite{Lim2014A}.
Eduardo proposed a Gabor filter to detect crack types \cite{EduardoZalama11}. Prasanna et al. presented spatially tuned robust multi-feature (STRUM) classifier  for crack detection \cite{Prasanna2016Automated}.

In early years, crack detection methods are mainly based on traditional image processing algorithms, which are divided into three steps: preprocessing,  preliminary detection, and refinement of cracks. Baltazart et al.\cite{kaul2011detecting} applied Minimal Path Selection algorithm without any prior knowledge to detect cracks within gray pavement images, which can extract the continuous cracks. Lim et al. \cite{lim2014robotic} used edge detection method for crack detection based on robotic crack inspection system.
However, traditional methods for crack detection are time-consuming and expensive.

In recent years, automatic methods were applied to accelerate crack detection and reduce costs for road managers and users. With the development of computer vision, deep learning has  entered the field of vision of researchers. Convolutional neural networks (CNNs) have achieved state-of-the-art performance in various computer vision applications.

Encoder-decoder framework with an U-shape is a typical backbone for segmentation tasks \cite{zou2018deepcrack}. Fan et al. proposed a U-HDN \cite{fan2020automatic} method based on encoder-decode architecture, including the proposed Multi-Dilation Module and Hierarchical Feature Module. These modules can obtain rich global context information and integrate multi-scale feature information, which can improve crack detection accuracy. Liu et al. \cite{liu2019deepcrack} proposed the DeepCrack for crack detection, which adopted the VGG16 neural network and hierarchical feature learning to improve neural network performance. However, it may fail to extract very thin cracks with complex topology.

Although many methods adopt the encoder-decoder framework to detect cracks and obtain  satisfactory results [4,24,27,35,36], these models are usually open-loop (OL) systems that fail to detect the thin cracks, and tend to treat them as the background with a good loss.

In 2014, Goodfellow et al. \cite{goodfellow2014generative} proposed the generative adversarial networks (GAN), which is widely used in machine learning and image generation tasks. GAN consists of two parts: a generative and an adversarial network. The generative one is used to generate fake images, and the latter distinguishes between real and generated results. After an alternative training, the generative network produces an image that resembles a real image, which can hardly be distinguished by the adversarial network. Inspired by this method, Zhang et al. \cite{zhang2020crackgan} proposed CrackGAN model to synthesise truth crack image, which uses crack patch supervised GAN and U-shape architecture to detect cracks. However, this method ignores the global feature of crack image, and could fail to detect complex road textures and obtain a relatively low accuracy for CFD dataset. Zhou et al. proposed Unet++ \cite{zhou2019unet++} to address the issue of the  global feature extraction of crack image using multiple skip connection operation based on U-net architecture. Although Unet++ indirectly integrates the features of different receptive fields, it only integrates the information of the next layer, and the information of the upper layer is not integrated. As a result, the granularity of its decoder part is still not fine enough, resulting in the loss of edge and position information in the segmentation results.
Gao et al. \cite{gao2019generative} modified the U-net with a different way of cross-layer concatenate ways and combined the discriminative network to perform crack segmentation. However, the output of the discriminative network adopts the two-class networks, which may simply divide the output patch image into a fake or real image, ignore the relationship between pixels, and cannot extract thin cracks with complex topology.

To tackle above problems, we embed closed-loop feedback (CLF) into the neural network so that the model could learn to correct errors on its own, which is called CrackCLF, including the front and back ends (i.e., segmentation and adversarial network, respectively). The proposed CrackCLF can perform end-to-end training based on generative adversarial networks (GAN) (Figure 1).

\begin{figure}
  \centering
  \includegraphics[width=8cm]{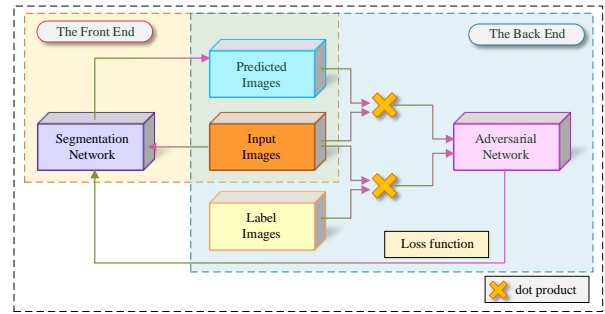}\\
  \caption{The proposed CrackCLF framework. The CrackCLF architecture contains two parts: Segmentation (the front end) and Adversarial network (the back end).}
  \label{CrackGAN framework}
\end{figure}

The proposed segmentation network (i.e., the front end) adopts an encoder-decoder architecture to perform crack segmentation (the dashed box on the left in Figure 1). In addition, we propose an upsampling-convolutional block attention module (UCBAM) to replace the convolution operation based on the decoder part, which can fuse features from the encoder and decoder parts. An innovative convolutional block attention module+ (CBAM+) being a lightweight model is embedded into UCBAM, which can reduce the number of parameters and calculations, and accelerate neural network training. Then, we designed the global attention pooling (GAP) based on CBAM+, which can assign different weights for the feature channel to focus on the crack information. We developed a hierarchical feature learning method to perform deep supervision of the crack output for accelerating the convergence of neural network.

The adversarial network (the back end) is employed to enforce the segmentation and adversarial network to learn global, local and thin crack information, which can capture a range of spatial relationships between pixels (the dashed box on the right in Figure 1). Meanwhile, it can correct higher-order inconsistencies between labels and crack maps (generated by the front end) to address open-loop system issues. In the training process of CrackCLF, the segmentation and adversarial networks are trained in an alternating manner.
The contributions of CrackCLF are:

\begin{enumerate}
\item 	A new crack detection framework, called CrackCLF, was proposed with a closed loop feedback, in which the front end is employed to segment crack, and the back end is used to distinguish images between the generated ones by the front end and the real images. To the best of our knowledge, this is the first time that a closed-loop system is integrated with  a mixed domain attention to perform crack detection task.

  \item  An upsampling-convolutional block attention module (UCBAM) is proposed to construct the decoder part in the proposed segmentation network (the front end), which can reduce the number of parameters and calculations costs, accelerating neural network training speed. In particular, a global attention pooling (GAP) is designed and embedded into the proposed CBAM+, which is part of UCBAM and can assign different weights for the feature channel to focus on the crack information.

  \item A hierarchical feature learning module is carefully designed and embedded into the segmentation network to perform deep supervision of the crack output for accelerating convergence of the neural network.

  \item An adversarial network (the back end) is proposed to perform closed-loop feedback to capture the difference between the predicted cracks and the ground truth, which can better learn features of global and local cracks to improve the performance of segmentation network.
\end{enumerate}

This study is organized as follows: In section II, we review the related works on crack detection. We describe the details of CrackCLF, including the segmentation and adversarial networks in Section III. Then, we perform comprehensive experiments to demonstrate the performance of CrackCLF in Section IV. Finally, Sections V is the conclusion of our study.

\section{Related Works}

\subsection{ Traditional Methods}

In 2006, Subirats et al. \cite{subirats2006automation} adopted a wavelet transform to detect cracks, based on different frequency sub-bands and amplitudes, which cannot be adapted to detect various types of cracks. Other researchers employed the threshold method to detect pavement cracks from images \cite{tang2013automatic,li2008novel}, followed by morphological operations to refine the cracks. Oliveira et al. \cite{oliveira2012automatic}  proposed CrackIT toolbox with threshold and edge detection methods et al., which is convenient for workers to detect crack images \cite{oliveira2014crackit}. However, edge detection, morphology and thresholding methods are sensitive to the background noises, such as oil stains, shadows, and leaves, which reduce the accuracy of crack detection. Minimal path-based methods consider brightness and connectivity to reduce noises and improve the accuracy of the continuous crack. Kaul et al. \cite{kaul2011detecting} adopted the minimal path selection (MPS) method without any prior knowledge, which can extract the continuous cracks. However, it fails to detect discontinuous cracks. In summary, the traditional methods normally need to adopt several steps (such as, eliminating noises and adjusting hyperparameters et al.) to detect cracks. Therefore, these methods are too computationally intensive for practical applications.

\subsection{Artificial Intelligence Methods}

To address the above issues, some researchers have proposed the use of artificial intelligence \cite{adeli2001neural}\cite{adeli1999machine123} to analyze cracks in road pavements. In particular, machine learning (ML) and deep learning (DL) have been proposed to perform semantic segmentation tasks \cite{tedeschi2017real}. Shi et al. \cite{shi2016automatic} proposed a random structured forests method with the designed feature descriptors and public pavement dataset (CFD) to help road managers to analyze and evaluate the pavement surfaces. However, this method heavily relied on the selection of feature descriptors cannot be generalized for different pavement types.  In general, the ML method cannot deal with and represent the features of the different pavement types for crack detection tasks.
In recent years, DL is widely used to perform crack detection tasks.

Zhang et al. \cite{zhang2016road} used crack patch images to train a convolutional neural network (CNN) to detect cracks on road pavements. However, the patch-based methods are usually very time-consuming and ignore the global crack information, resulting in relatively low accuracy. In CrackNet, Zhang et al. \cite{zhang2017automated,zhang2019automated} adopted a CNN without pooling layers to avoid crack information loss, improve the pixel-perfect accuracy, and achieve a higher speed in crack detection. Fan et al. developed a structured prediction and an ensemble network method with the patch-based to extract cracks \cite{fan2018automatic} and make measurements \cite{fan2020ensemble}. However, these methods may ignore the global crack feature and can fail to extract complex and thin cracks. Meanwhile when these methods are used to train large-scale datasets for crack detection, they obtain a lower accuracy with smaller convolution layers.

Encoder-decoder framework with an U-shape is a typical backbone for segmentation tasks \cite{zou2018deepcrack}. Fan et al. proposed a U-HDN \cite{fan2020automatic} method based on encoder-decode architecture, including the proposed Multi-Dilation Module and Hierarchical Feature Module. These modules can obtain rich global context information and integrate multi-scale feature information, which can improve crack detection accuracy.  Qu et al. \cite{qu2021deeply} proposed a new method based on deeply supervised convolutional neural network for pavement crack detection with Multi-Scale Features Fusion, which adopted U framework  and hierarchical feature learning to improve neural network performance. Guo et al. \cite{guo2021barnet} proposed BARNet method, which includes three modules: Edge Adaptation Module, Base Predictor Module and Refinement Module. Knig \cite{Knig2023} proposed using the  Weakly-supervised method to segmentation crack image, which can obtain a better accuracy than others. Sun et al.\cite{sun2022dma} proposed a DMA (DeepLab With Multi-Scale Attention) method based on DeepLab, which integrates the ASPP(Atrous Spatial Pyramid Pooling) and attention mechanism. Zhou et al.\cite{zhou2022method} proposed an Enhanced Convolution and Dynamic Feature Fusion (ECDFFNet) method, which can improve its overall performance by focusing on the local details.

\section{Methods}
According to the literature in this study, the segmentation network is employed to detect cracks, and the adversarial one is used to perform feedback operation and enforce the segmentation to learn global, local and thin crack information.
\subsection{Segmentation Network (the Front End)}

The encoder-decoder architecture is used in the segmentation network, which consists of three parts: an encoder part, a decoder part, and a hierarchical feature learning module.

\textbf{(1) Encoder Part:}

In the encoder part, each encoder structure consists of two 3 $\times$ 3 convolutional layers and activation function (ReLU) \cite{nair2010rectified}. Meanwhile, the pad operation is embedded into the 3 $\times$ 3 convolution operation to maintain the original resolution. Then, a pooling layer is adopted to downsampling operation to reduce the image size. After these operations, the image size is halved, and the number of channels is doubled, such as 64, 128, 256, 512, and 1024 (Figure 2).

\begin{figure*}
  \centering
  \includegraphics[width=14cm]{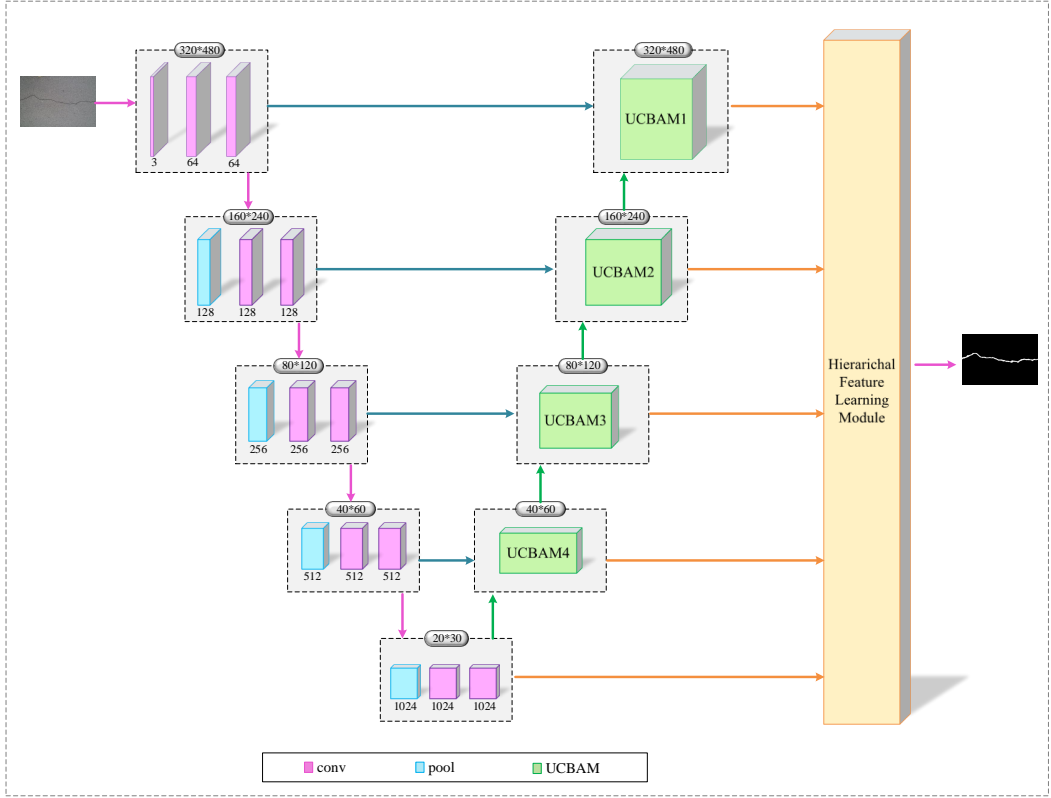}\\
  \caption{The proposed Segmentation Network framework (the front end). The Segmentation Network contains three parts: encoder part, decoder part and hierarchical feature learning module.}
  \label{Segmentation_Network framework}
\end{figure*}

\textbf{(2) Decoder Parts:}

The decoder part is composed of 4 successive UCBAM modules to restore the image size. Finally, the number of channels is gradually reduced to 1 (Figure 2).

\subsubsection{ \textbf{The Proposed UCBAM}}

The resolution of the feature maps was reduced after passing through many downsampling operations in the final part of the encoder. The feature maps in the encoder part have more information and are blurred and discarded after the downsampling operation. If we extract the crack image with a simple upsampling operation from the encoder part, the extracted crack results can be rough. To perform crack segmentation, the size of the final features may be restored to the original size. The function of the decoder is to restore the size of the features using transposed convolution operation. To fuse and incorporate more information, the skip connection from the encoder features is used to connect the decoder part. However, if these two different feature maps are simply added, the different contributions of edge and texture cannot be presented in the segmentation process. Hence, to overcome this issue, restore the image size, and highlight the important semantic information, UCBAM is proposed to perform channel and spatial attention mapping (Figure 3).

The inputs are the features from the encoder part with a skip connection along with the previous decoder part, and the output features are optimized after weight fusion. UCBAM first restores the size of the images from the previous decoder part using transposed convolution (light blue cuboid in Figure 3). Then, it adds the features (yellow cuboid in Figure 3) from the encoder part to generate the fused feature maps (pink cuboid in Figure 3), which can obtain and highlight the important semantic information, such as edges and textures.

In Figure 3 the yellow cuboid represents the feature maps from the encoder part by skip connection. The light blue cuboid is used to restore the image size that is from the previous decoder part after transpose convolution. Then, these two types of features perform an addition operation to fuse feature maps. Finally, the fused feature maps (pink cuboid) are input into the proposed CBAM+ module to obtain weighted features. Subsequently, the proposed CBAM+ based on CBAM \cite{woo2018cbam} is employed to use the fused features to learn the weights of different features and suppress unnecessary features. UCBAM can adaptively assign different weights to the channels which can learn and highlight the different crack semantic information.

Specifically, the feature maps-based decoder part performs transposed convolution operations to restore the resolution and reduce the number of channels. The modified features are then concatenate to features from encoder to generate fused features. Finally, the fused features are given as input to the proposed CBAM+ to generate weight feature maps. The proposed UCBAM is able to reduce parameters¡¯ number and calculations costs and accelerate the neural network training speed.

\begin{figure*}
  \centering
  \includegraphics[width=14cm]{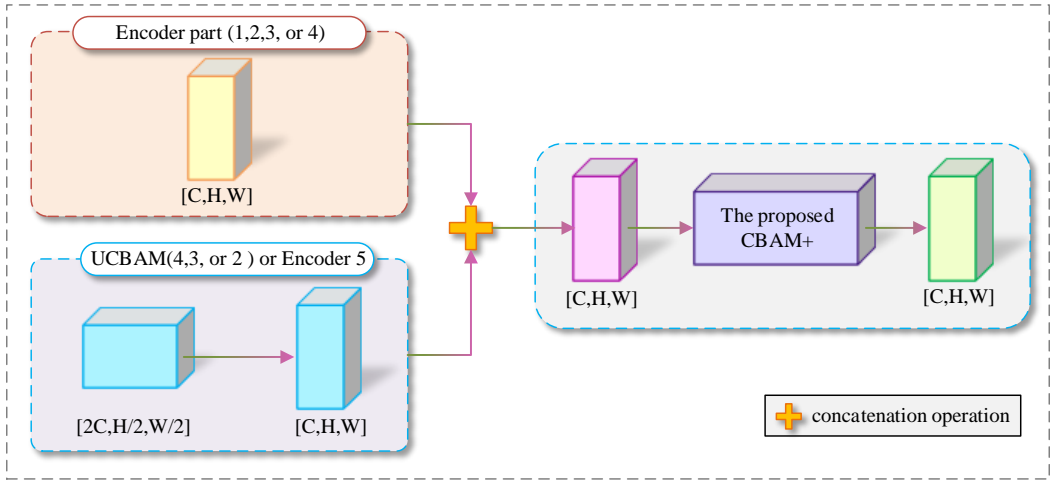}\\
  \caption{The proposed UCBAM module framework. The number of the feature maps¡¯ channel, height, and width are [C, H, W], respectively.}
  \label{UCBAM framework}
\end{figure*}

\subsubsection{ \textbf{The Proposed CBAM+}}

The attention mechanism is widely embedded into the neural network to enhance the performance of models, reduce training time, and make the neural network lightweight. In CBAM \cite{woo2018cbam}, channel and spatial attention are employed to focus on significant features and ignore redundant features. This combination is superior to channel attention. Meanwhile, the CBAM is only able to obtain local features in the convolution process, which cannot capture and extract long-range dependency.

To overcome this issue, the CBAM+ module is proposed to capture the global context and make the neural network lightweight in this study. It contains channel attention and spatial attention modules (Figure 4). Specifically, the channel attention (CA) module extracts long-range dependency, capture the global edges, patterns, textures and enhance the performance of the model, and the spatial attention (SA) module was employed based on CBAM \cite{woo2018cbam}. CBAM+ embedded into the UCBAM can improve calculation speed for the decoder part.

\begin{figure*}
  \centering
  \includegraphics[width=14cm]{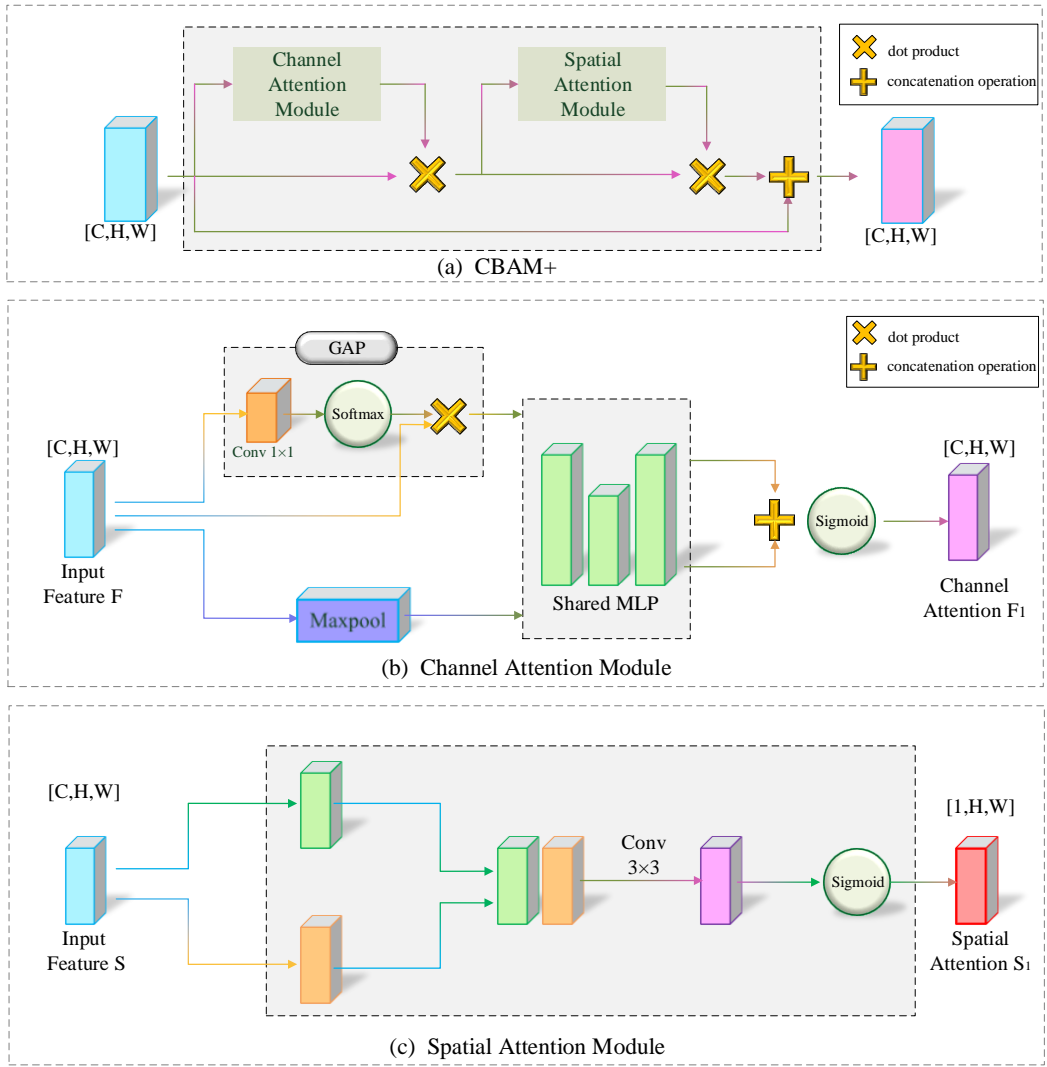}\\
  \caption{Diagram of the proposed CBAM+ module. }
  \label{CBAM+_channel_spatial framework}
\end{figure*}

\textbf{The Proposed Channel Attention Module:}
We propose a Channel Attention (CA) module to replace the original one based on CBAM \cite{woo2018cbam}. CA can exploit the inter-channel relationship of features and focus on the important information, given an input image. The designed global attention pooling (GAP) performs pooling operations (Figure 4(a)), which can generate the attention weights to be assigned to global context features. Equation 1 gives the output of the GAP ($F_{gap}$):

\begin{equation}\label{global_attention_pooling}
 F_{gap}=\sum \limits_{j=1}^{N_p}\frac{e^{W_k \cdot x_j}}{\sum \limits_{m=1}^{N_p}{e}^{W_k \cdot x_m}}x_j
\end{equation}

where $x$ is defined as an input feature, $W_k$ is the linear transform matrix (1$\times$1 convolution operation), $N_p$ is the number of pixels (positions), and $x_j$ enumerates all possible pixels (positions). $F_{gap}$ is the output of the context modeling. Meanwhile, $\alpha_j=\frac{e^{W_k \cdot x_j}}{\sum \limits_{m=1}^{N_p}{e}^{W_k \cdot x_m}}$ is defined as the weight of the global attention pooling. Specifically, the global attention pooling is passed through a 1$\times$1 convolution, followed by a softmax function to generate attention weights, which can assign different weights to the features to obtain the context feature maps and capture long-range dependency.

On the other hand, the max pooling ($F_{max}$) gives significant information about desired features to infer finer channel attention. $F_{gap}$ and $F_{max}$ are employed to perform the channel attention feature extraction. Subsequently, the two pooling results are passed through a shared network to produce the proposed CA feature maps (Equation 2).

\begin{equation}\label{tondaojizhi}
\begin{aligned}
  F_1 &=\sigma (Conv(GAPooling(F)) + Conv (MaxPooling(F))) \\
  &= \sigma (W_1(W_0(F_{gap}))+ W_1(W_0(F_{max})))
\end{aligned}
\end{equation}
where $F_1$ and $\sigma$ denote the outputs of the channel attention maps and sigmoid function, respectively. $W_0 \in \mathbb{R}^{C/r \times C}$, $W_1 \in \mathbb{R}^{C \times C/r}$, and $Conv$ is a $1 \times 1$ convolution operation.

\textbf{Spatial Attention Module:} In this study, we used the SA module based on CBAM \cite{woo2018cbam} to produce a spatial attention map that captures the inter-spatial relationship of features. Compared with CA, this module focuses on the position of the region containing important information (Equation 3). The average and max pooling ($S_{avg}$ and $S_{max}$, respectively) are employed to perform spatial operations to produce the SA maps with a convolution. The spatial attention module is represented by equation (3):

\begin{equation}\label{kognjianzhuyili}
\begin{aligned}
  S_1 &=\sigma(Conv([AvgPooling(S);MaxPooling(S)])) \\
  &= \sigma(Conv([S_{avg};S_{max}]))
\end{aligned}
\end{equation}
where $S$ is the input feature and AvgPooling and Maxpooling are defined as the average pooling and max pooling, respectively. Average pooling features: $S_{avg} \in\mathbb{R}^{1 \times H \times W } $, max pooling features: $S_{max} \in\mathbb{R}^{1 \times H \times W } $. $Conv$ is a $3 \times 3$ convolution operation.

\textbf{(3) Hierarchical feature-learning module:}

The different-level features include the different complex context, patterns, and textures information. Therefore, we adopted the hierarchical feature learning module to extract crack for obtaining the different crack semantic information individually (Figure 5). The designed hierarchical feature learning module can accelerate the convergence of neural network with deeply supervised nets (DSN) \cite{lee2015deeply}.

\begin{figure}
  \centering
  \includegraphics[width=8cm, height=10cm]{9hierarchical_feature_learning.pdf}\\
  \caption{The hierarchical feature learning module framework.}
  \label{hierarchical_feature_learning}
\end{figure}
Each side output and fused output are supervised by DSN \cite{lee2015deeply} according to the holistically-nested edge detection network (HED)  \cite{xie2015holistically}. A training dataset is defined as $S=(X_n,Y_n), n=1, ..., N$, where $X_n$ and $Y_n$ are the raw original image and ground truth crack map, respectively. Each side network is followed by a classifier, and the weights for each side network are denoted as $w=(w^{(1)}, ..., w^{(M)})$.  $W$ and $M$ are the number of network parameters and side networks, respectively. Equation (4) represents the loss function for the side networks.
\begin{equation}
  L_{side}(W,w)=\sum\limits_{m=1}^{M} \alpha_{m} l^{m}_{side}(W,w^{m})
\end{equation}
where $l_{side}$ and  $\alpha_m$ are the loss function and a weight hyperparameter at each side. The $l_{side}$ loss function was applied to distinguish the non-crack and crack pixels with  equation (5):
\begin{equation}
  l_{side}=\frac{1}{N}\sum\limits_{i=1}^{N}\{\beta y_i \log \hat{y}_i +\gamma(1- y_i)\log(1-\hat{y}_i)\}
\end{equation}
where $\beta$ and $\gamma$ are hyperparameters, and $N$ is defined as the number of pixels in one image. $y_i$ and $\hat{y}_i$ are the label and predicted result, respectively.

The entitle outputs of the side networks are concatenated to produce fused feature maps (5 channels), and the fused loss function $L_{fuse}$ is equation (6):
\begin{equation}
  L_{fuse}=\frac{1}{N}\sum\limits_{i=1}^{N}\{\beta y_i \log \hat{y}_i +\gamma(1- y_i)\log(1-\hat{y}_i)\}
\end{equation}
where $\beta$ and $\gamma$ have the same meanings as in Equation (5). Finally, the total loss function is defined according to  equation (7):
\begin{equation}
  L_{total}=L_{side}+L_{fuse}
\end{equation}

\begin{figure*}
  \centering
  \includegraphics[width=14cm]{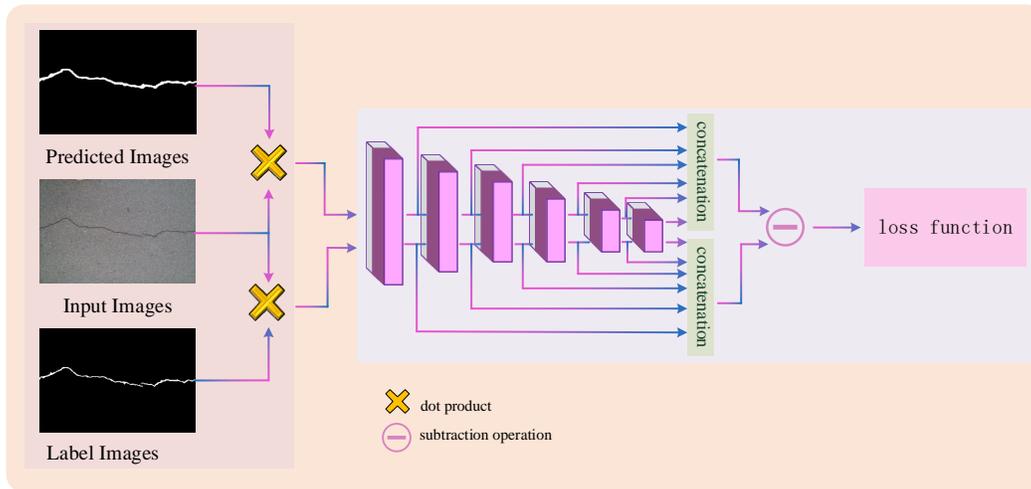}\\
  \caption{An adversarial network framework (the back end). An adversarial network adopts the multi-scale $L_1$ loss to solve the mapping relationship between input images and generated segmentation images.}
  \label{adversarial_network framework}
\end{figure*}
\subsection{Adversarial Network (the Back End)}

Unlike traditional GAN, the adversarial network developed in this study is based on the multi-scale $L_1$ loss to deal with mapping relationship between input and generated segmentation images (Figure 6) to perform closed-loop feedback for  correcting high-order inconsistencies between ground truth and the predicted result and improve the detection accuracy of unrecognized small cracks in the segmentation network.

The input follows two paths (Figure 6): the former is the result of the predicted images being multiplied by the input images and the latter is the result of multiplying the labelled images by the input images. The multi-scale object loss function is employed to calculate the mean absolute error operation for features between two paths (Equation 6) and transfer to segmentation network with backpropagation operation (Figure 1), which can improve the detection accuracy of unrecognized small cracks, and correct high-order inconsistencies between ground truth and the predicted result in the training process.

An adversarial network adopts the multi-scale $L_1$ loss function based on pixels level to solve the mapping relationship between input images and generated segmentation images. The multi-scale $L_1$ loss function is demonstrated with equation (8):

\begin{equation}\label{duochidusunshihanshu}
  \min \limits_{\theta_S} \max\limits_{\theta_C}L_1(\theta_S,\theta_C) = \frac{1}{N}\sum \limits_{n=1}^{N} \ell_{mae}(f_C(x_n\circ S(x_n)),f_C(x_n \circ y_n))
\end{equation}
where $N$ is the number of images, and $x_n$ and $y_n$ are the input image and corresponding label image, respectively. $\ell_{mae}$ denotes the mean absolute error. $x_n\circ S(x_n) $ is defined as the result of multiplying the predicted image by input image. $x_n \circ y_n$ denotes the results obtained by multiplying the label image by the input image. $f_C(x)$  is a hierarchical feature extracted from image $x$ by the adversarial network.  $\ell_{mae}$ can be represented by  equation (9):
\begin{equation}\label{l_mae_function}
 \ell_{mae}(f_C(x),f_C(x^{'}))=\frac{1}{N}\sum \limits_{i=1}^{N}\| f^{i}_C(x)-f^{i}_C(x^{'})\|_{1}
\end{equation}
where $N$ is the number of layers in the adversarial network, and $ f^{i}_C(x)$ is the extracted feature at the $i$th layer based on the adversarial network.

The multi-scale $L_1$ loss can capture long- and short- spatial relations between pixels by using the different level features, i.e., low-, middle- and high-level features, which can perform feedback operation (closed-loop feedback) to enforce segmentation and capture the difference between the predicted cracks and the ground truth and improve the accuracy of the segmentation network.

\section{Experiments and Results}

Firstly, we mainly introduce experimental details of the CrackCLF in this section. Next, we present the evaluation metrics and compare them. Finally, we analyze and demonstrate experimental results.

\subsection{Implementation Details}

The proposed CrackCLF method was programmed using the Pytorch library \cite{paszke2017automatic} as the deep learning framework for training and testing based on the GPU server with four NVIDIA TITAN Xp GPU, each having 12GB of memory.

\subsubsection{Crack Dataset}
The public databases CFD \cite{shi2016automatic} were used to evaluate the CrackCLF. The CFD database contains 118 images (resolution 320 $\times$ 480).  72 images and 46 images were used to train and test the proposed CrackCLF, respectively.

The Crack500 dataset \cite{yang2019feature} contains 500 images of size about $2000 \times1500$, which were collected by phones in the main campus of the Temple University. The dataset is cropped into 16 non-overlapping image regions. Finally, training, validation, and test data contained 1896, 346, and 1124 images. All images share the same size of $256\times256$ in the training, validation, and test phases.

The Crack700 dataset was collected from Harbin, China \cite{yang2018automatic}, which contains 776 different types of images, such as concrete walls and bridge surfaces. The images were taken at different distances, which led to different resolutions. All images share the same size of $256\times256$ in the training, validation, and test processes.

\subsubsection{Evaluate metrics}
The precision (Pr), recall (Re), and F1 score (F1) were used to evaluate CrackCLF using the following equations:
\begin{equation}
  Pr= \frac{TP}{TP+FP}
\end{equation}
\begin{equation}
  Re= \frac{TP}{TP+FN}
\end{equation}
\begin{equation}
  F1= \frac{2\times Pr \times Re}{Pr+Re}
\end{equation}
where $TP, FP$, and $FN$ are the number of true positives, false positives, and false negatives, respectively. $F1$ was employed to evaluate the overall performance of crack detection. Specifically, two different metrics based on $F1$ were adopted:  the best $F1$ on the public database for a fixed threshold (ODS), and the aggregate $F1$ on the public database for the best threshold in each image (OIS).

ODS and OIS are defined as the max value according to Equations 13 and 14, respectively.
\begin{equation}
  ODS=\frac{2\times Pr_t \times Re_t}{Pr_t+Re_t}: t=0.001,...0.999
\end{equation}
\begin{equation}
  OIS=\frac{1}{N_{img}}\sum _{i}^{N_{img}} max{\frac{2\times Pr_t^{i} \times Re_t^{i}}{Pr_t^{i}+Re_t^{i}}: t=0.001,...0.999}
\end{equation}
where $t, i$, and $N_{img}$ are the threshold, index, and number of images, respectively. $Pr_t, Re_t, Pr_t^{i}$, and $Re_t^{i}$ are the precision and recall based on the threshold $t$ and image $i$, respectively.

We consider the transitional areas between the non-crack and crack pixels before computing $TP, FP$, and $FN$. 2 pixels distances are also accepted in this study \cite{amhaz2016automatic,fan2018automatic,ai2018automatic,konig2019convolutional,fan2020ensemble}. The decision threshold is defined as 0.5. The hyperparameter contains: batch size (4 images for CFD, 16 images for Crack 500 and Crack700), selecting adam optimzer, epochs 500, learning rate (0.001).

\subsection{Experimental Results and Analysis}
In this section, we mainly compare different algorithms based on different datasets to demonstrate the performance of the proposed CrackCLF method.

\subsubsection{CFD dataset}

\begin{figure*} 
  \centering
  \includegraphics[height=8cm, width=16cm]{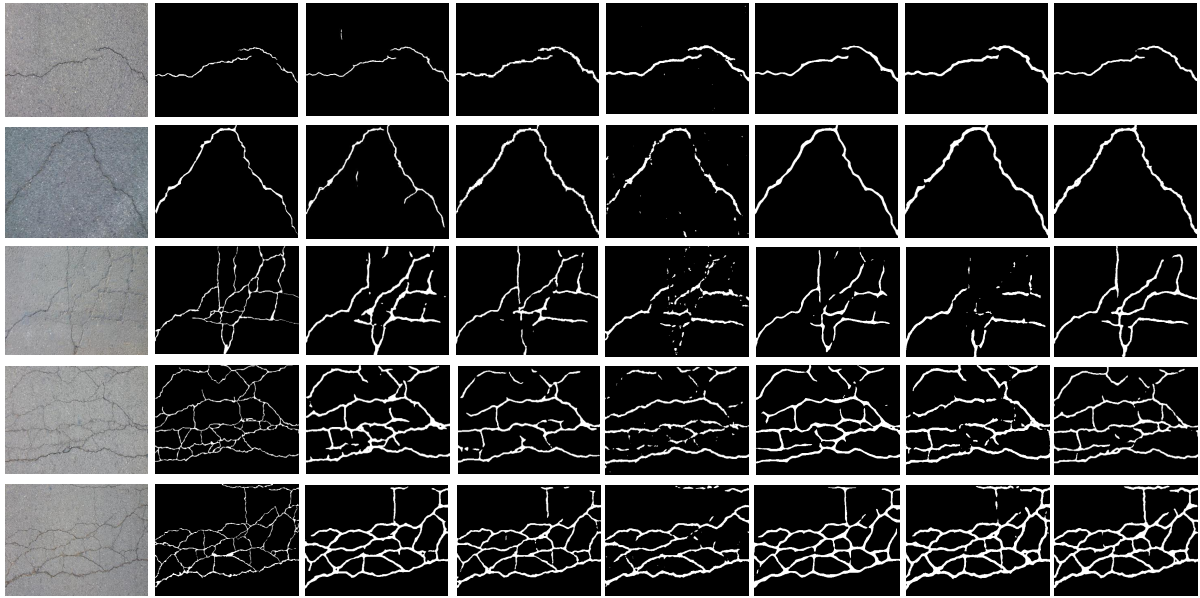}
  \caption{Experimental results of comparison of the proposed CrackCLF with other methods based on CFD (from left to right: 1) original image, 2) groundTruth,  3) U-net, 4) U-HDN algorithm, 5) Segnet, 6) DeepCrackZou, 7) DeepCrackLiu,  8) CrackCLF).}
  \label{fig:CFD_output_crackgan1}
\end{figure*}

Table \ref{tab:cfd-comparison} lists the results in terms of Pr, Re, and F1 for different methods on CFD. Bold characters refer to the highest values of each column. The local thresholding method \cite{oliveira2009automatic} is sensitive to noise and unable to inspect cracks (Pr equals to 0.7727 and F1 equals to 0.7418). The CrackForest method \cite{shi2016automatic} was able to extract the wider cracks than the ground truth, which leads to low Pr (0.74466) and high Re (0.9514) values: it can extract many non-crack areas. As is shown in Figure \ref{fig:CFD_output_crackgan1}, it was observed that  U-net \cite{ronneberger2015u} and SegNet \cite{badrinarayanan2017segnet} can extract the crack skeleton, but there are some misidentifications in the images (Pr equals to 0.9119, 0.9325, 0.9256, 0.8876, respectively). U-HDN \cite{fan2020automatic}, DeepCrackLiu \cite{liu2019deepcrack}, and DeepCrackZou \cite{zou2018deepcrack} adopt the U-shape framework method to perform crack detection based on hierarchical feature learning, which can extract the crack skeleton. Although CrackGAN\cite{zhang2020crackgan} has a larger recall, it has a lower precision and F1, which can overestimate the crack regions with lower accuracy (Figure \ref{fig:CFD_output_crackgan1}).

However, traditional methods consider only pixel-to-pixel comparisons in the prediction phase, while CrackCLF transforms the feedback from the adversarial to the segmentation network improving the detection accuracy (Pr equals to 0.9451) and ensuring satisfactory performance. The F1 values demonstrate that CrackCLF gave the best overall performance of the crack detection (i.e., 0.9406) and the highest Pr value (i.e., 0.9451).

Table \ref{tab:cfd_ods_ois} lists the ODS and OIS values of competing methods based on CFD. Bold characters refer to the highest values of each column. Compared with other methods, CrackCLF gave the best performances in terms of ODS (0.9374) and OIS (0.9455). So, the CrackCLF can extract cracks information based on global and local databases (Figure \ref{fig:CFD_output_crackgan1}).

\begin{table}
  \centering
  \caption{Crack detection experimental results on CFD dataset.}
    \begin{tabular}{cccc}
    \toprule
    \textbf{Method}  & \textbf{Pr} & \textbf{Re} & \textbf{F1} \\
    \midrule
    Canny\cite{zhao2010improvement}      & 0.4377 & 0.7307 & 0.457 \\
    Local thresholding \cite{oliveira2009automatic}     & 0.7727 & 0.8274 & 0.7418 \\
    CrackForest \cite{shi2016automatic}      & 0.7466 & 0.9514 & 0.8318 \\
    CrackSeg \cite{gao2019generative}     & 0.7618 & 0.5959 & 0.6413 \\
    MFCD \cite{li2018automatic}     & 0.899 & 0.8947 & 0.8804 \\
    Method \cite{ai2018automatic}      & 0.907 & 0.846 & 0.87 \\
    Structured prediction \cite{fan2018automatic}      & 0.9119 & 0.9481 & 0.9244 \\
    Ensemble network \cite{fan2020ensemble}     & 0.9256 & \textbf{0.9611} & 0.934 \\
    U-net \cite{ronneberger2015u}      & 0.9325 & 0.932 & 0.928 \\
    U-HDN \cite{fan2020automatic}      & 0.945 & 0.936 & 0.938 \\
    U-net ++ \cite{zhou2019unet++}      & 0.9412 & 0.935 & 0.938 \\
    DSMS \cite{zhou2022method} & 0.9422 & 0.9388 & 0.941 \\
    DMA \cite{sun2022dma} & 0.9435 & 0.9331 & 0.9382 \\
    BARNET \cite{guo2021barnet} & 0.9333 & 0.941 & 0.9356 \\
    Segnet \cite{badrinarayanan2017segnet}     & 0.8876 & 0.9073 & 0.8937 \\
    DeepCrack£¨Zou£©\cite{zou2018deepcrack}      & 0.9447 & 0.9276 & 0.9364 \\
    DeepCrack£¨Liu£©\cite{liu2019deepcrack}     & 0.912 & 0.936 & 0.92 \\
    CrackGAN\cite{zhang2020crackgan}	&0.8803	&\textbf{0.9611}	&0.9189 \\
    CrackCLF    & \textbf{ 0.9451 }& 0.9344 & \textbf{0.9406} \\
    \bottomrule
    \end{tabular}%
  \label{tab:cfd-comparison}%
\end{table}%
\begin{table}
  \centering
  \caption{The ODS and OIS of competing methods on CFD dataset.}
    \begin{tabular}{ccc}
    \toprule
    \textbf{Methods} & \textbf{ODS} & \textbf{OIS} \\
    \midrule
    HED  \cite{xie2015holistically} & 0.593 & 0.626 \\
    RCF  \cite{liu2017richer} & 0.542 & 0.607 \\
    FCN  \cite{long2015fully} & 0.585 & 0.609 \\
    CrackForest \cite{shi2016automatic} & 0.104 & 0.104 \\
    FPHBN \cite{yang2019feature} & 0.683 & 0.705 \\
    U-net \cite{ronneberger2015u} & 0.923 & 0.931 \\
    U-net ++ \cite{zhou2019unet++}      & 0.933 & 0.943  \\
    DSMS \cite{zhou2022method} & 0.936  & 0.942 \\
   DMA \cite{sun2022dma} & 0.932  & 0.9422 \\
    BARNET \cite{guo2021barnet} & 0.9362 & 0.9435 \\
    U-HDN \cite{fan2020automatic} & 0.935 & 0.928 \\
    Segnet \cite{badrinarayanan2017segnet} & 0.8938 & 0.9091 \\
    DeepCrack(Zou)\cite{zou2018deepcrack} & 0.9305 & 0.94 \\
    DeepCrack(Liu)\cite{liu2019deepcrack} & 0.9242 & 0.937 \\
    CrackCLF & \textbf{0.9374} & \textbf{0.9455} \\
    \bottomrule
    \end{tabular}%
  \label{tab:cfd_ods_ois}%
\end{table}%

\subsubsection{Crack500 dataset}

\begin{figure*} 
  \centering
  \includegraphics[height=15cm,width=15cm]{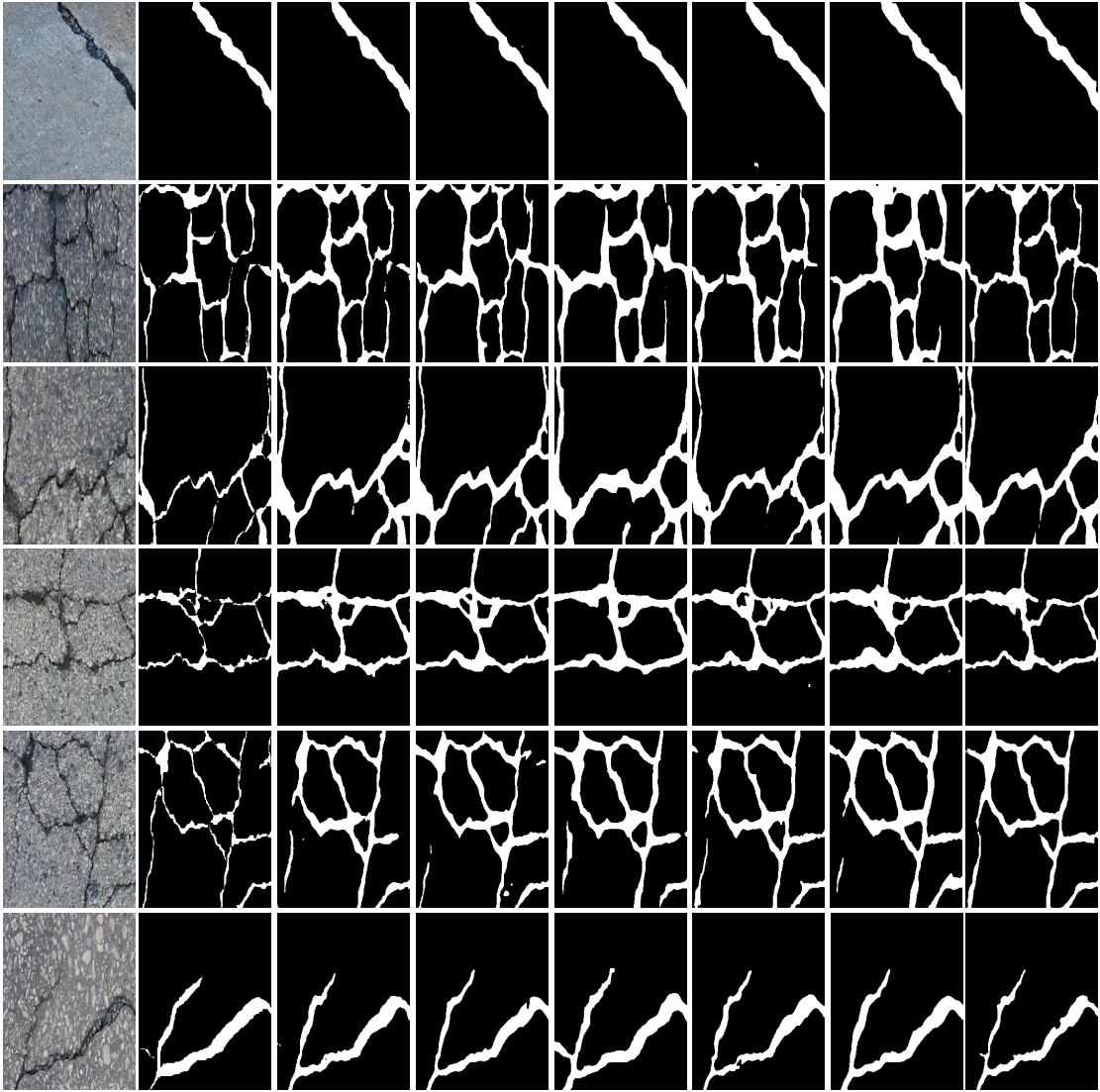}
  \caption{Experimental results of comparison of the proposed CrackCLF with other methods based on Crack500 (from left to right: 1) original image, 2) groundTruth, 3) Segnet, 4) U-net, 5) DeepCrackLiu, 6) U-HDN algorithm, 7) DeepCrackZou, 8) CrackCLF).}
  \label{fig:crack500_output_crackgan}
\end{figure*}

Figure \ref{fig:crack500_output_crackgan}, Table \ref{tab:crack500-1} and Table \ref{tab:crack500-2} present some examples of detection on Crack500. Bold characters refer to the highest values of each column. Segnet \cite{badrinarayanan2017segnet} and DeepCrackLiu \cite{liu2019deepcrack} were not able to recognize and detect thinner cracks (Pr equal to 0.7591 and 0.7661, respectively), and thereby unable to extract consecutive cracks (F1 equal to 0.7855 and 0.7885, respectively). Based on large Re (0.8632) and low Pr (0.7526), U-net \cite{ronneberger2015u} can not obtain the satisfactory results because it can overestimate and extract non-crack areas: it can inspect the thinner cracks, but it overestimates the crack width, which can lead to lower accuracy (Figure \ref{fig:crack500_output_crackgan}).

U-HDN \cite{fan2020automatic} can inspect thinner cracks, but discontinuous cracks also occur in the image as is showing in Figure \ref{fig:crack500_output_crackgan}. Furthermore, it was observed that some isolated pixels were recognized as cracks. DeepCrackZou \cite{zou2018deepcrack} cannot detect thin cracks and overestimates the crack width, which can cause high Re (0.9064) and low Pr (0.6655) values. CrackCLF can recognize and inspect the thinner crack and extract the continuous crack pixels obtaining satisfactory performance: compared with other methods, CrackCLF gave the best performances in terms of Pr (0.7793) and OIS (0.8225).

In summary, it is clear that CrackCLF can obtain two best performance results (Pr:0.7793, F1:0.7913) than others, and CrackCLF can obtain satisfactory accuracy. Pr metrics is higher than second method U-HDN (Pr:0.7744). In term of ODS and OIS, the proposed method obtain the best OIS (0.8225) and the fourth ODS (0.8096). It is obvious that the CrackCLF can get best accuracy in each image for different threshold value t with the OIS equation. Although the ODS is not the best result, the CrackCLF can obtain three best results (Pr, F1, and OIS) in five metrics. Therefore, CrackCLF ensures satisfactory performances.
\begin{table}[htbp]
  \centering
  \caption{Crack detection experimental results on Crack500 dataset.}
    \begin{tabular}{cccc}
    \toprule
    \textbf{Method}  & \textbf{Pr} & \textbf{Re} & \textbf{F1} \\
    \midrule
    Structured prediction \cite{fan2018automatic} & 0.3867 & 0.7984 & 0.4751 \\
    Ensemble network \cite{fan2020ensemble}& 0.4147 & 0.7763 & 0.4953 \\
    SegNet \cite{badrinarayanan2017segnet} & 0.7591 & 0.8504 & 0.7855 \\
    U-net \cite{ronneberger2015u} & 0.7526 & 0.8632 & 0.7864 \\
    U-net ++ \cite{zhou2019unet++}      & 0.7625 & 0.853 & 0.7872 \\
    DSMS \cite{zhou2022method} & 0.7616 & 0.8322 & 0.7885 \\
    DMA \cite{sun2022dma} & 0.765 & 0.813 & 0.784 \\
    BARNET \cite{guo2021barnet} & 0.7733 & 0.831 & 0.7862 \\
    DeepCrack£¨Zou£©\cite{zou2018deepcrack} & 0.6655 & \textbf{0.9064} & 0.7499 \\
    DeepCrack£¨Liu£©\cite{liu2019deepcrack} & 0.7661 & 0.8503 & 0.7885 \\
    U-HDN \cite{fan2020automatic} & 0.7744 & 0.8234 & 0.7788 \\
    CrackCLF & \textbf{0.7793} & 0.8404 & \textbf{0.7913} \\
    \bottomrule
    \end{tabular}%
  \label{tab:crack500-1}%
\end{table}%

\begin{table}[htbp]
  \centering
  \caption{The ODS and OIS of competing methods on Crack500 dataset.}
    \begin{tabular}{ccc}
    \toprule
    \textbf{Method} & \textbf{ODS} & \textbf{OIS} \\
    \midrule
    HED \cite{xie2015holistically}  & 0.575 & 0.625 \\
    RCF \cite{liu2017richer}  & 0.49  & 0.586 \\
    FCN  \cite{long2015fully}  & 0.513 & 0.577 \\
    CrackForest \cite{shi2016automatic} & 0.199 & 0.199 \\
    FPHBN \cite{yang2019feature} & 0.604 & 0.635 \\
    SegNet \cite{badrinarayanan2017segnet} & 0.8116 & 0.8117 \\
    U-net \cite{ronneberger2015u} & 0.8128 & 0.8131 \\
    U-net ++ \cite{zhou2019unet++}      & 0.8148 & 0.8188  \\
    DSMS \cite{zhou2022method} & 0.813  & 0.8191 \\
    DMA \cite{sun2022dma} & 0.818  & 0.821 \\
    BARNET \cite{guo2021barnet} & 0.8193 & 0.8211 \\
    DeepCrack£¨Zou£©\cite{zou2018deepcrack} & 0.8023 & 0.8155 \\
    DeepCrack£¨Liu£©\cite{liu2019deepcrack} & \textbf{0.8197} & 0.819 \\
    U-HDN  \cite{fan2020automatic} & 0.807 & 0.8091 \\
    CrackCLF & 0.8096 & \textbf{0.8225} \\
    \bottomrule
    \end{tabular}%
  \label{tab:crack500-2}%
\end{table}%

\subsubsection{Crack700 dataset}

\begin{figure*} 
  \centering
  \includegraphics[height=14cm, width=13cm]{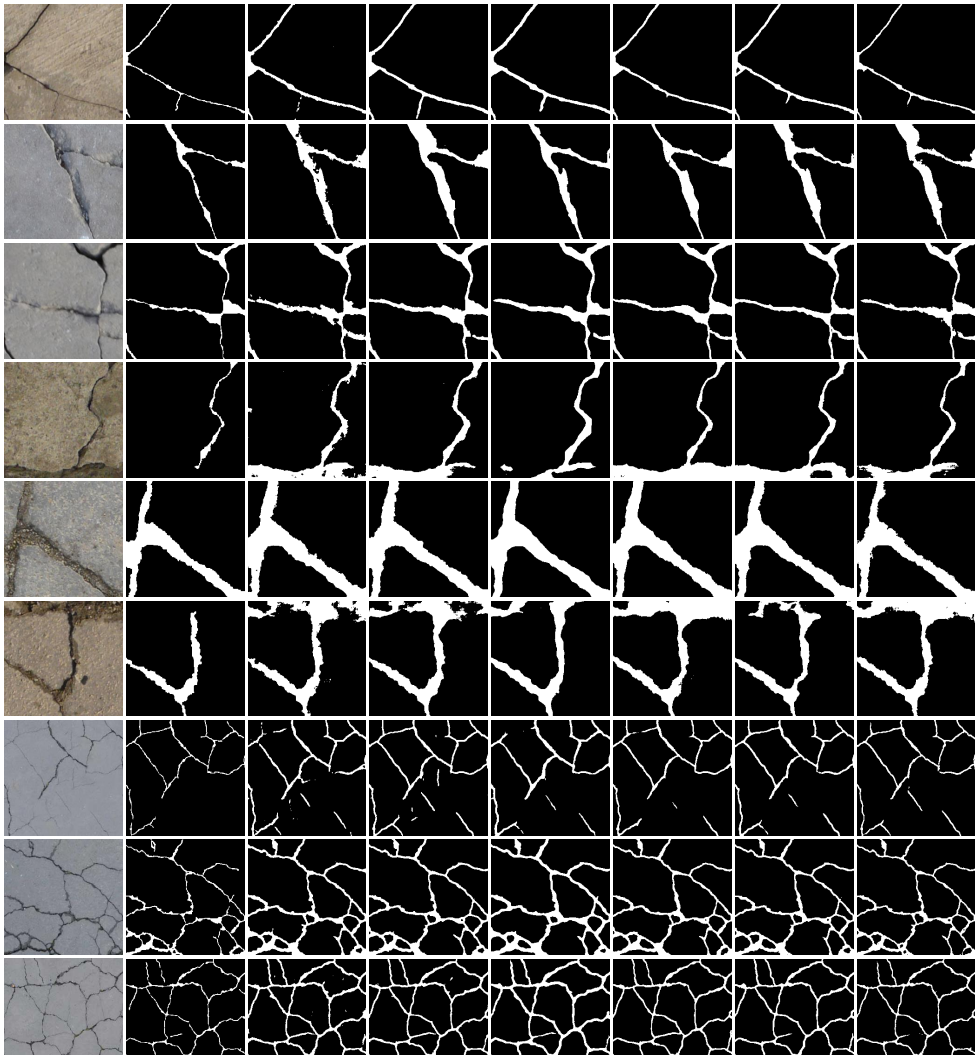}
  \caption{Experimental results of comparison of the proposed CrackCLF with other methods based on Crack700 (from left to right: 1) original image, 2) groundTruth, 3) Segnet, 4) U-net, 5) DeepCrackLiu, 6) DeepCrackZou, 7) U-HDN, 8) CrackCLF).}
  \label{fig:crack700}
\end{figure*}

Figure \ref{fig:crack700}, Table \ref{tab:Crack700pingjiazhibiao-1} and Table \ref{tab:Crack700pingjiazhibiao-2} list some examples of detection on Crack700. Bold characters refer to the highest values of each column. Under such conditions, Segnet \cite{badrinarayanan2017segnet} was able to extract the skeleton architect of the crack, but there were some isolated holes pixels of non-crack and crack. U-net \cite{ronneberger2015u}, DeepCrackLiu \cite{liu2019deepcrack}, DeepCrackZou \cite{zou2018deepcrack}, and U-HDN \cite{fan2020automatic} algorithms were able to extract the crack skeleton, but they tend to overestimate the crack width, which can lead to high Re (0.9760, 0.9677, 0.9533, and 0.9550, respectively) and low Pr (0.8670, 0.8756, 0.9008, and 0.9040, respectively) values. Compared with other algorithms, CrackCLF gave low Re (0.9457), the highest Pr (0.9151) and the highest F1 (0.9237) because it can extract thin and consecutive crack pixels.

From Table \ref{tab:Crack700pingjiazhibiao-2}, it is clear that the proposed CrackCLF can obtain the best ODS (0.9181) and the second OIS (0.9364, only small 0.0004). It presents that CrackCLF can extract more crack information for different thresholds based on the global database with equation 13. In short, the CrackCLF get three best results (Pr:0.9151, F1:0.9237, and ODS:0.9182) than others based on five metrics.

\begin{table}[htbp]
  \centering
  \caption{Crack detection experimental results based on Crack700 dataset.}
    \begin{tabular}{cccc}
    \toprule
    \textbf{Method} & \textbf{Pr} & \textbf{Re} & \textbf{F1} \\
    \midrule
    Structuted prediction \cite{fan2018automatic} & 0.8217 & 0.9317 & 0.8584 \\
    Ensemble network \cite{fan2020ensemble} & 0.827 & 0.9385 & 0.8631 \\
    U-HDN \cite{fan2020automatic} & 0.904 & 0.955 & 0.922 \\
    U-net \cite{ronneberger2015u} & 0.867 & \textbf{0.976} & 0.9101 \\
    U-net ++ \cite{zhou2019unet++}      & 0.8825 & 0.953 & 0.9172 \\
    DSMS \cite{zhou2022method} & 0.9027 & 0.9327 & 0.9185 \\
    DMA \cite{sun2022dma} & 0.9078 & 0.9344 & 0.9211 \\
    BARNET \cite{guo2021barnet} & 0.9033 & 0.9331 & 0.9206 \\
    Segnet \cite{badrinarayanan2017segnet} & 0.8511 & 0.9635 & 0.8934 \\
    DeepCrackLiu \cite{liu2019deepcrack}& 0.8756 & 0.9677 & 0.9105 \\
    DeepCrackZou \cite{zou2018deepcrack}& 0.9008 & 0.9533 & 0.9187 \\
    CrackCLF & \textbf{0.9151} & 0.9457 & \textbf{0.9237} \\
    \bottomrule
    \end{tabular}%
  \label{tab:Crack700pingjiazhibiao-1}%
\end{table}%

\begin{table}[htbp]
  \centering
  \caption{The ODS and OIS of competing methods on Crack700 dataset.}
    \begin{tabular}{ccc}
    \toprule
    \textbf{Method} & \textbf{ODS} & \textbf{OIS} \\
    \midrule
    U-HDN \cite{fan2020automatic} & 0.916 & 0.934 \\
    U-net\cite{ronneberger2015u} & 0.911 & 0.933 \\
    U-net ++ \cite{zhou2019unet++}      & 0.916 & 0.935 \\
    DSMS \cite{zhou2022method} & 0.913  & 0.931 \\
    DMA \cite{sun2022dma} & 0.912  & 0.935 \\
    BARNET \cite{guo2021barnet} & 0.917 & 0.933 \\
    Segnet \cite{badrinarayanan2017segnet} & 0.892 & 0.918 \\
    DeepCrackLiu \cite{liu2019deepcrack} & 0.9151 & \textbf{0.9368} \\
    DeepCrackZou \cite{zou2018deepcrack} & 0.909 & 0.929 \\
    CrackCLF & \textbf{0.9182} & 0.9364 \\
    \bottomrule
    \end{tabular}%
  \label{tab:Crack700pingjiazhibiao-2}%
\end{table}%
\subsection{Ablation Study}

To demonstrate the superiority of the CrackCLF framework, ablation study is adopted to verify the performance of the CrackCLF and the advantages of the different modules from different crack datasets (Table \ref{tab:ablation-study}) with the encoder U-net. CBAM, CBAM+, the adversarial network, and the hierarchal feature learning modules.

CBAM+ could perform better accuracy (larger Pr, F1, ODS and OIS for CFD) than CBAM, and CBAM gives a higher Re than CBAM+. When the adversarial network was embedded into CrackCLF, Pr and F1 were enhanced (Pr equal to 0.9459 for CFD with adversarial network, 0.7793 for Crack500 with adversarial and hierarchical network, and 0.9191 for Crack700 with adversarial network, showing the adversarial network can improve the detection accuracy; F1 equal to 0.9604 for CFD with adversarial and hierarchical network, 0.7913 for Crack500 with adversarial and hierarchical network, and 0.9237 for Crack700 with adversarial network), and Re was reduced (Re equal to 0.9337 for CFD, 0.8346 for Crack500, and 0.9289 for Crack700 with adversarial network).

The proposed CBAM+ embedded into the UCBAM can not only exploit the inter-channel relationship of features and focus on the important information and enhance the accuracy for different databases from Table VII, but also reduce the parameters' number and calculations costs, accelerate the neural network training speed.

Meanwhile, some metrics¡¯ values are further to enhancing and the neural network can extract thinner cracks by the embedded adversarial network with closed-loop feedback for correcting high-order inconsistencies between the label and the predicted image with segmentation network. Finally, the hierarchical feature module with a deep-supervised network and fusing feature maps accelerates the convergence speed and improves model performance.

\begin{figure*}
  \centering
  \includegraphics[width=15cm]{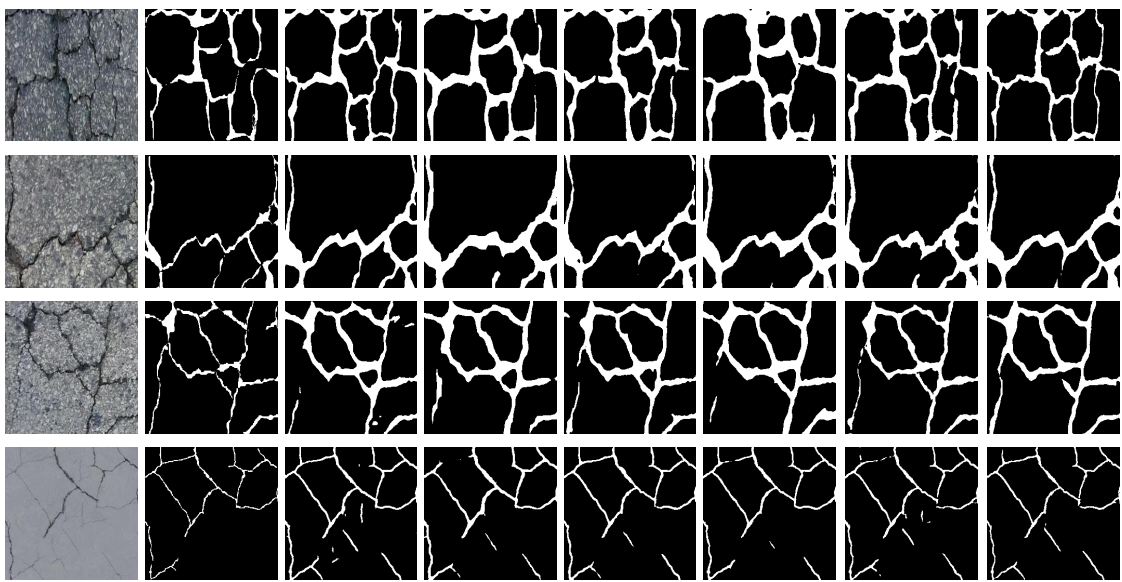}
  \caption{Comparison of the CrackCLF with or without adversarial network (from left to right: 1) original image, 2) Groundtruth,
  3) U-net, 4) DeepCrackLiu, 5) U-HDN, 6) DeepCrackZou, 7) CrackCLF without adversarial network, 8) CrackCLF).}
  \label{fig:thincrack}
\end{figure*}

Figure \ref{fig:thincrack} shows the comparison of the proposed method with existing methods in terms of detecting thinner cracks. U-net method can overestimate the crack features, as is shown in Figure \ref{fig:thincrack}, which result in a low accuracy. In this example, DeepCrack(Liu) cannot extract the continuous shallow cracks very well, which results in a low precision. U-HDN method was not able to very well detect continuous shallow cracks either and fail to extract some thin cracks. DeepCrack(Zou) method tends to extract the wider cracks and ignore the continuous thin ones, which can cause the higher recall and lower accuracy. Figure \ref{fig:thincrack} also shows that the CrackCLF without the adversarial network cannot extract the thin cracks well, which leads to lower precision. Meanwhile, it can be seen that the CrackCLF without adversarial network overestimates the crack regions, causing larger recall and lower precision, as showing in Table \ref{tab:ablation-study}. The proposed CrackCLF with closed-loop feedback can better extract continuous thin crack features, which effectively improves the accuracy of crack detection. Meanwhile, the width for crack feature is better identified by our method in comparison with other methods. In summary, the proposed CrackCLF with closed-loop feedback performs better in extracting shallow cracks.

\begin{table*}[htbp]
  \centering
  \caption{The comparison of ablation study based on different modules.}
    \begin{tabular}{ccccccccccc}
    \toprule
    \textbf{Dataset} & \textbf{Encoder} & \textbf{CBAM} & \textbf{CBAM+} & \textbf{Adversarial Network} & \textbf{Hierarichal Feature } & \textbf{Pr} & \textbf{Re} & \textbf{F1} & \textbf{ODS} & \textbf{OIS} \\
    \midrule
    \multirow{4}[2]{*}{\textbf{CFD}} & \multicolumn{1}{c}{\multirow{12}[6]{*}{\textbf{U-net\newline{} Encoder}}} & \checkmark     &       &       &       & 0.9178 & \textbf{0.9509} & 0.9321 & 0.9343 & 0.9411 \\
          &       &       & \checkmark     &       &       & 0.9335 & 0.9441 & 0.937 & 0.9367 & 0.9421 \\
          &       &       & \checkmark     & \checkmark     &       & \textbf{0.9459} & 0.9337 & 0.9379 & 0.9351 & 0.9432 \\
          &       &       & \checkmark     & \checkmark     & \checkmark     & 0.9451 & 0.9344 & \textbf{0.9406} & \textbf{0.9374} & \textbf{0.9455} \\
\cmidrule{1-1}\cmidrule{3-11}    \multirow{4}[2]{*}{\textbf{Crack500}} &       & \checkmark     &       &       &       & 0.7222 & \textbf{0.8755} & 0.7729 & 0.7997 & 0.8064 \\
          &       &       & \checkmark     &       &       & 0.7299 & 0.874 & 0.7776 & 0.8016 & 0.8092 \\
          &       &       & \checkmark     & \checkmark     &       & 0.7756 & 0.8346 & 0.7861 & 0.8055 & 0.8163 \\
          &       &       & \checkmark     & \checkmark     & \checkmark     & \textbf{0.7793} & 0.8404 & \textbf{0.7913} & \textbf{0.8096} & \textbf{0.8225} \\
\cmidrule{1-1}\cmidrule{3-11}    \multirow{4}[2]{*}{\textbf{Crack700}} &       & \checkmark     &       &       &       & 0.9048 & 0.9471 & 0.9174 & 0.9125 & 0.9292 \\
          &       &       & \checkmark     &       &       & 0.9061 & \textbf{0.9479} & 0.9198 & 0.9111 & 0.9314 \\
          &       &       & \checkmark     & \checkmark     &       & \textbf{0.9191} & 0.9289 & 0.9159 & 0.9124 & 0.9338 \\
          &       &       & \checkmark     & \checkmark     & \checkmark     & 0.9151 & 0.9457 & \textbf{0.9237} & \textbf{0.9182} & \textbf{0.9364} \\
    \bottomrule
    \end{tabular}%
  \label{tab:ablation-study}%
\end{table*}%

\subsection{Model Complexity}

We utilize the number of the parameters (Params), floating-point operations per second (FLOPs) and frames per second (FPS) to evaluate the processing efficiency of CrackCLF and other models (Table \ref{tab:cfd_flops_params}).
It can be observed from the results that compared with other methods, the proposed CrackCLF achieves satisfactory efficiency based on the experimental results although FLOPs and Params are not the smallest (Table \ref{tab:cfd_flops_params}). However, the costs of calculation for the proposed CrackCLF only with the segmentation network (FLOP:17.02G) are smallest among the compared ones, which can run pretty fast (30FPS) to detect crack images during the inference stage with relatively small Params (18.84M). Thanks to the fewest FLOPs and least parameter number, DeepCrack(Liu) and HED achieve the fastest inference speed of 33 FPS (0.03 second per image), and can detect cracks in real time (normally, 24 FPS for human eye). Our method obtains a running speed of 30 FPS, exactly 0.033 second per image, also achieving a real-time detection. This is because in practice, only the segmentation network (the front end) is employed at the crack inference stage after training is finished, and the adversarial network (the back end) is not needed in real world deployment. In short, the proposed CrackCLF is able to obtain superior detection accuracy with a satisfactory detection speed, which can meet the requirements of most real-world application scenarios.

\begin{table}[htbp]
  \centering
  \caption{The comparison of model complexity.}
    \begin{tabular}{cccc}
    \toprule
    \textbf{Method} & \textbf{FLOPs} & \textbf{Params} & \textbf{FPS(F/s)} \\
    \midrule
    HED \cite{xie2015holistically}  & 20.07G & \textbf{14.72M} & \textbf{33} \\
    RCF \cite{liu2017richer}  & 20.44G & 14.80M & 33 \\
    U-net \cite{ronneberger2015u} & 65.47G & 34.53M & 13 \\
    Segnet \cite{badrinarayanan2017segnet} & 40.16G & 29.44M & 16 \\
    DeepCrack£¨Zou£©\cite{zou2018deepcrack} & 42.12G & 31.78M & 16 \\
    DeepCrack£¨Liu£©\cite{liu2019deepcrack} & 20.11G & \textbf{14.72M} & \textbf{33} \\
    U-HDN \cite{fan2020automatic} & 38.66G & 29.12M & 16 \\
    CrackGAN \cite{zhang2020crackgan} &51.54G & 34.77M  & 13\\
    Unet++ \cite{zhou2019unet++} &107.78G & 43.23M  & 6\\
    CrackCLF(only Segmentation) & \textbf{17.02G} & 18.84M & 30 \\
    CrackCLF & 26.04G & 26.39M & 18 \\
    \bottomrule
    \end{tabular}%
  \label{tab:cfd_flops_params}%
\end{table}%

\subsection{Visualization for  UCBAM Modules}

\begin{figure*}
  \centering
  \includegraphics[width=14cm]{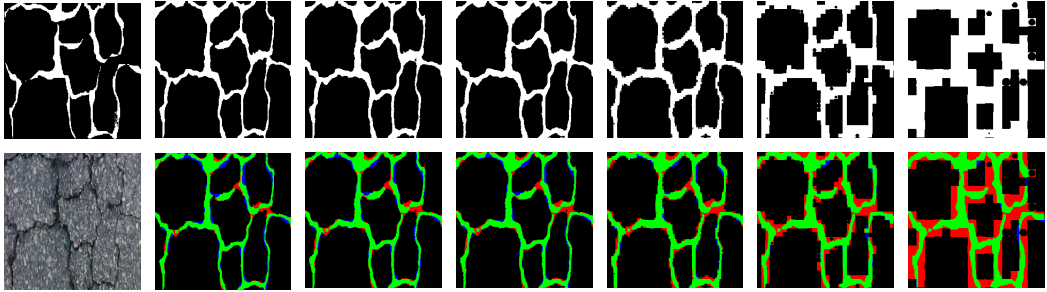}
  \caption{Visualization of feature maps for each UCBAM module. The green, red, and blue pixels represent true positives (TP), false positives (FP), and false negatives (FN), respectively. (from left to right: 1) Original Image or Groundtruth, 2) Output, 3) UCBAM 1, 4) UCBAM 2, 5) UCBAM 3, 6) UCBAM 4, 7) Encoder 5).}
  \label{fig:UCBAMoutput}
\end{figure*}

The proposed  CBAM+ module can exploit the inter-channel relationship of features and focus on the important information, given an input image. It can improve the network¡¯ ability to extract high- and low- level semantic information, which is crucial for accurate crack segmentation. In Figure \ref{fig:UCBAMoutput}, we visualize the feature maps UCBAM modules and encoder 5, to intuitively observe the effect of these modules in the proposed network. After being processed by these modules, the feature maps are clearly enhanced on the crack areas. Based on the enhanced processed features, the classifier can more readily recognize the crack areas and perform an even more accurate crack segmentation.

Meanwhile, this design enables the module to consider both high-level and low-level characteristics in determining the weights of channels and screening channels that better characterize cracks. By visualizing the output of the UCBAM module, we can clearly observe that the discrimination between crack and non-crack pixels is greatly facilitated by focusing on the significantly reduced regions highlighted by the proposed attention mechanism.

\subsection{Generalization Capability of CLF Module}

It is also worthwhile to point out that one of the most important contributions of this work is that we propose a CLF mechanism that can be integrated into almost any current models as a plug-in module, and further improve their performances by introducing a close loop feedback to address the open-loop issue of existing models.

In order to validate this, we conducted a comprehensive ablation study, in which we compared many existing models with and without the CLF mechanism, as shown in Table \ref{tab:addlabeCLF}. It can be observed that the state-of-the-art methods without CLF tend to have a higher Re, but lower Pr and F1 score than their counterparts with CLF mechanism, which means that integrating CLF can effectively improve their performances. In particular, the back end (CLF) is used to correct higher-order inconsistencies between labels and crack maps (generated by the front end) to address the issue of open-loop systems, which can help improve the performance of the networks. In a word, the proposed CLF can be defined as a plug and play module, which can be embedded into different neural network models to improve their performances.

\begin{table}[htbp]
  \centering
  \caption{Crack detection experimental results based on three datasets. ¡¯w/¡¯ and ¡¯w/o¡¯ mean ¡¯with¡¯ and ¡¯without¡¯, respectively.}
    \begin{tabular}{ccccc}
    \toprule
    \textbf{Datasets} & \textbf{Methods} & \textbf{Pr} & \textbf{Re} & \textbf{F1} \\
    \midrule
    \multirow{12}[2]{*}{\textbf{CFD}} & U-net w/ CLF \cite{ronneberger2015u} & \textbf{0.9345} & 0.9315 & \textbf{0.9326} \\
          & U-net w/o CLF \cite{ronneberger2015u} & 0.9325 & \textbf{0.932} & 0.928 \\
          & U-net ++ w/ CLF  \cite{zhou2019unet++} & \textbf{0.9457} & 0.9287 & \textbf{0.9391} \\
          & U-net ++ w/o CLF  \cite{zhou2019unet++} & 0.9412 & \textbf{0.935} & 0.938 \\
          & DeepCrackZou w/ CLF \cite{zou2018deepcrack} & \textbf{0.9452} & 0.9256 & \textbf{0.9377} \\
          & DeepCrackZou w/o CLF \cite{zou2018deepcrack} & 0.9447 & \textbf{0.9276} & 0.9364 \\
          & DeepCrackLiu w/ CLF \cite{liu2019deepcrack} & \textbf{0.9266} & 0.9355 & \textbf{0.9299} \\
          & DeepCrackLiu w/o CLF \cite{liu2019deepcrack} & 0.912 & \textbf{0.936} & 0.92 \\
          & U-HDN w/ CLF \cite{fan2020automatic} & \textbf{0.9458} & \textbf{0.9378} & \textbf{0.9392} \\
          & U-HDN w/o CLF \cite{fan2020automatic} & 0.945 & 0.936 & 0.938 \\
          & CrackCLF w/ CLF & \textbf{0.9451} & 0.9344 & \textbf{0.9406} \\
          & CrackCLF w/o CLF & 0.9335 & \textbf{0.9441} & 0.937 \\
    \midrule
    \multirow{12}[2]{*}{\textbf{Crack500}} & U-net w/ CLF \cite{ronneberger2015u} & \textbf{0.7645} & 0.8533 & \textbf{0.7886} \\
          & U-net w/o CLF \cite{ronneberger2015u} & 0.7526 & \textbf{0.8632} & 0.7872 \\
          & U-net ++ w/ CLF  \cite{zhou2019unet++} & \textbf{0.7689} & 0.8466 & \textbf{0.7911} \\
          & U-net ++ w/o CLF  \cite{zhou2019unet++} & 0.7625 & \textbf{0.853} & 0.7872 \\
          & DeepCrackZou w/ CLF \cite{zou2018deepcrack} & \textbf{0.6956} & 0.8823 & \textbf{0.7612} \\
          & DeepCrackZou w/o CLF \cite{zou2018deepcrack} & 0.6655 & \textbf{0.9064} & 0.7499 \\
          & DeepCrackLiu w/ CLF \cite{liu2019deepcrack} & \textbf{0.7713} & 0.8423 & \textbf{0.7894} \\
          & DeepCrackLiu w/o CLF \cite{liu2019deepcrack} & 0.7661 & \textbf{0.8503} & 0.7885 \\
          & U-HDN w/ CLF \cite{fan2020automatic} & \textbf{0.7812} & 0.8123 & \textbf{0.7823} \\
          & U-HDN w/o CLF \cite{fan2020automatic} & 0.7741 & \textbf{0.8234} & 0.7788 \\
          & CrackCLF w/ CLF & \textbf{0.7793} & 0.8404 & \textbf{0.7913} \\
          & CrackCLF w/o CLF & 0.7299 & \textbf{0.874} & 0.7761 \\
    \midrule
    \multirow{12}[2]{*}{\textbf{Crack700}} & U-net w/ CLF \cite{ronneberger2015u} & \textbf{0.8811} & 0.9522 & \textbf{0.9156} \\
          & U-net w/o CLF \cite{ronneberger2015u} & 0.867 & \textbf{0.976} & 0.9101 \\
          & U-net ++ w/ CLF \cite{zhou2019unet++} & \textbf{0.8956} & 0.9423 & \textbf{0.9201} \\
          & U-net ++ w/o CLF \cite{zhou2019unet++} & 0.8825 & \textbf{0.953} & 0.9172 \\
          & DeepCrackZou w/ CLF \cite{zou2018deepcrack} & \textbf{0.9128} & 0.9425 & \textbf{0.9219} \\
          & DeepCrackZou w/o CLF \cite{zou2018deepcrack} & 0.9008 & \textbf{0.9533} & 0.9187 \\
          & DeepCrackLiu w/ CLF \cite{liu2019deepcrack} & \textbf{0.8934} & 0.9422 & \textbf{0.9188} \\
          & DeepCrackLiu w/o CLF \cite{liu2019deepcrack} & 0.8756 & \textbf{0.9677} & 0.9105 \\
          & U-HDN w/ CLF \cite{fan2020automatic} & \textbf{0.9149} & 0.9423 & \textbf{0.9256} \\
          & U-HDN w/o CLF \cite{fan2020automatic} & 0.904 & \textbf{0.955} & 0.922 \\
          & CrackCLF w/ CLF & \textbf{0.9151} & 0.9457 & \textbf{0.9237} \\
          & CrackCLF w/o CLF & 0.9061 & \textbf{0.9479} & 0.9198 \\
    \bottomrule
    \end{tabular}%
  \label{tab:addlabeCLF}%
\end{table}%

\section{Conclusion}
Automatic pavement crack detection is an imperative task to ensure functional and structural performances of road pavements. To tackle an open-loop (OL) system with encoder-decoder framework, we introduce closed-loop feedback (CLF) in the neural network so that the model could learn to correct errors on its own, based on generative adversarial networks (GAN), which is called CrackCLF and includes the front and back end (segmentation and adversarial network). The segmentation network contains two parts: encoder and decoder. In the encoder part, we employ the U-net encoder pat  and we propose the UCBAM module to replace the convolution operation in the decoder part. Meanwhile, the proposed CBAM+ module is embedded into the UCBAM module, and the designed hierarchical feature module is employed to obtain the multiscale feature maps of different decoder parts. An adversarial network is used to enforce the segmentation and adversarial networks to learn global and local crack information to overcome open-loop system issues. The proposed framework has been compared with other methods (i.e., Canny \cite{zhao2010improvement}, Local thresholding \cite{oliveira2009automatic}, CrackForest \cite{shi2016automatic}, MFCD \cite{li2018automatic}, Structured prediction \cite{fan2018automatic}, Ensemble network \cite{fan2020ensemble}, U-net \cite{ronneberger2015u}, U-HDN  \cite{fan2020automatic}, Segnet \cite{badrinarayanan2017segnet}, DeepCrackZou \cite{zou2018deepcrack}, DeepCrackLiu \cite{liu2019deepcrack} using three public datasets (i.e., CFD, Crack500, and Crack700). CrackCLF can give satisfactory output in terms of Re, Pr, and F1 (not less than 0.7793, 0.8404, and 0.7913, respectively). Finally, the ablation study verifies the performance of the CrackCLF with different modules based on different crack datasets. In summary, the proposed CrackCLF can give satisfactory results on three public datasets. Moreover, the proposed CLF can be defined as a plug and play module, which can be embedded into different neural network models to improve their performances.

In this study, CrackCLF gave promising results. However, there are still some limitations to be addressed in future work.
 \begin{itemize}
   \item Because the automatic crack detection system is only used to detect individual images so far. Video streaming will be tested in future studies.

   \item Moreover, the artificially designed neural network may contain redundant feature maps. Designing a neural network that can automatically optimize and prune its structure and parameters will be further investigated in our future work.

   \item The size of CrackCLF can be further reduced to become a more lightweight model, so that it can be more easily deployed in devices with limited computing resources, or adversary environments where computing capability of the devices will be largely restrained.
    \item The mechanisms of CrackCLF may be integrated with those of SAM to further improve its performance. This can be a new direction that is worthwhile to be investigated in the future.

 \end{itemize}


%



%
%

\ifCLASSOPTIONcaptionsoff
  \newpage
\fi



\bibliographystyle{IEEEtran}
\bibliography{references}

\begin{IEEEbiography}
[{\includegraphics[width=1in,height=1.15in,clip,keepaspectratio]{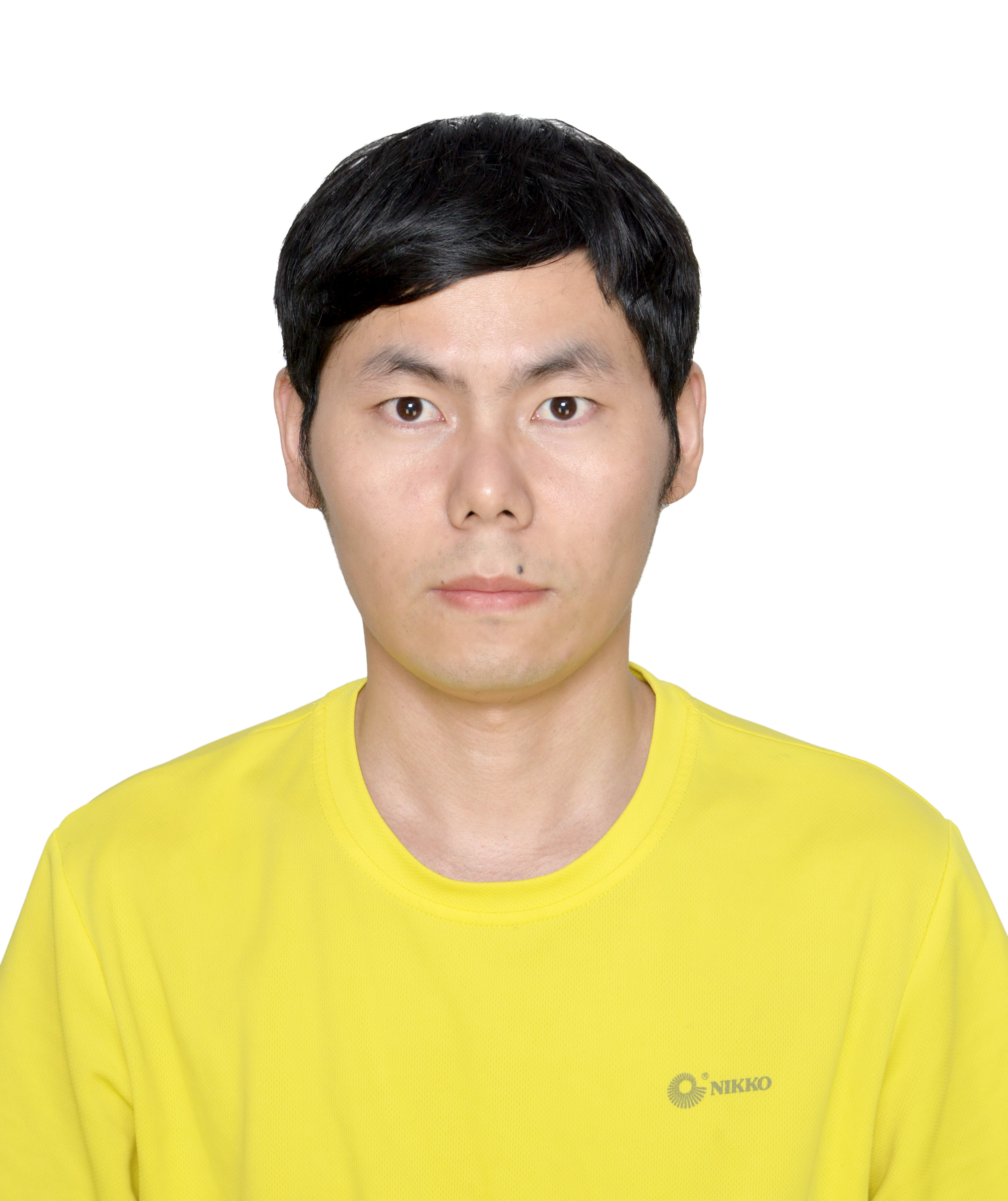}}]{Chong Li}
received the M.S. degree in electronic engineering from Shantou University, Shantou, China, in 2018. He is currently pursuing the Ph.D. degree with the Structural Engineering, Shantou University. His current research interests include image processing and machine learning.
\end{IEEEbiography}
\begin{IEEEbiography}
[{\includegraphics[width=1in,height=1.15in,clip,keepaspectratio]{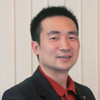}}]{Zhun Fan (Senior Member, IEEE)}
 received the B.S. and M.S. degrees in control engineering from the Huazhong University of Science and Technology, Wuhan, in 1995 and 2000, respectively, and the Ph.D. degree in electrical engineering from Michigan State University, East Lansing, USA, in 2004. He is currently a full Professor and Head of the Department of Electronic and Information Engineering, Shantou University. His research interests include artificial intelligence and computer vision.
\end{IEEEbiography}
\begin{IEEEbiography}
[{\includegraphics[width=1in,height=1.15in,clip,keepaspectratio]{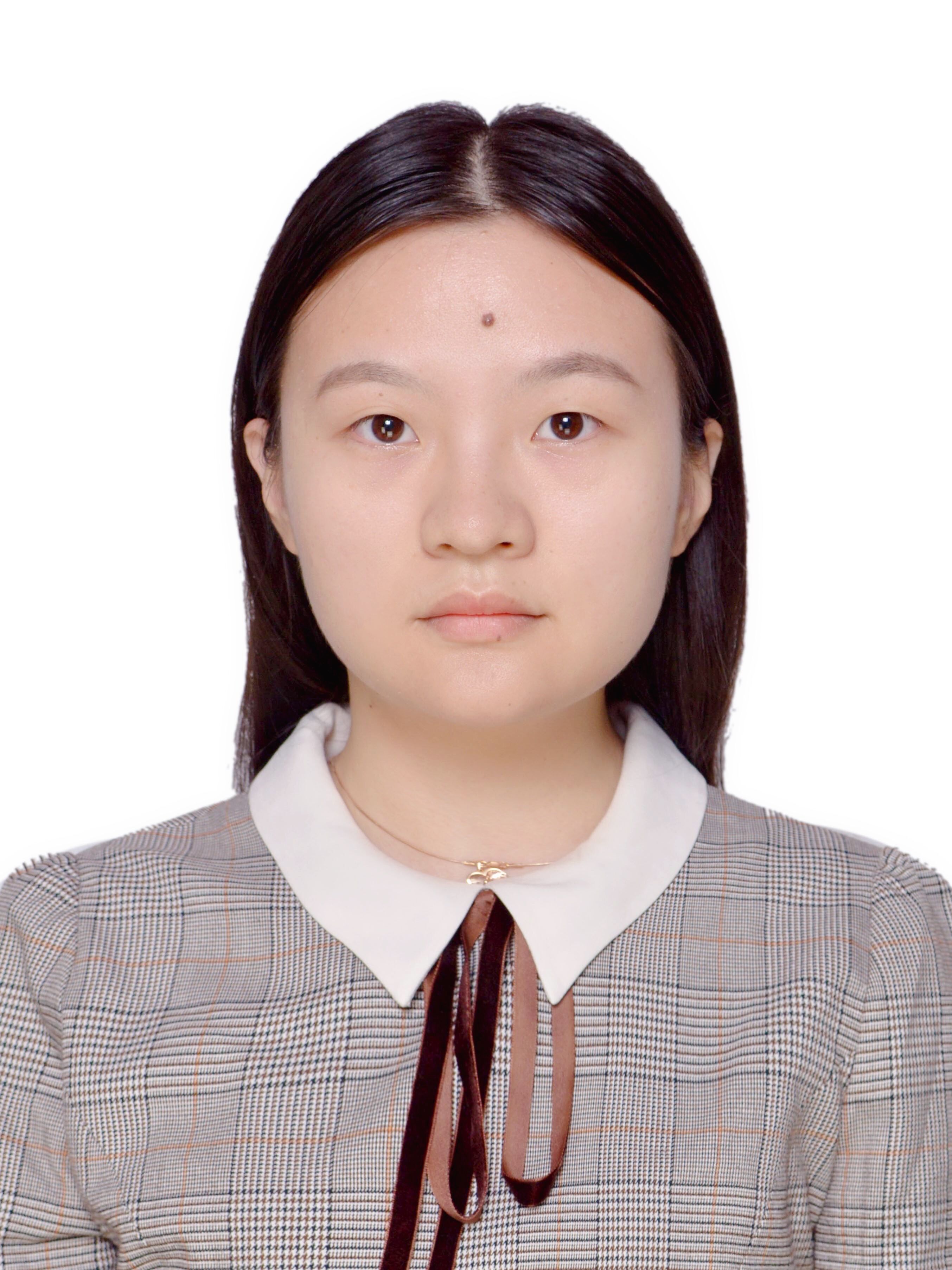}}]{Ying Chen}
received the M.S. degree in Shantou University, Shantou, in 2022. Her current research interests include medical image analysis, image processing, and deep learning.
\end{IEEEbiography}
\begin{IEEEbiography}
[{\includegraphics[width=1in,height=1.15in,clip,keepaspectratio]{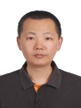}}]{Huibiao Lin}
received the M.S. degree in artillery automatic weapons and ammunition engineering from Shenyan ligong University, Shenyan, in 2008. He is currently pursuing the Ph.D. degree in structural engineering from the college of engineering, Shantou University, Shantou, China. His research interests include computational intelligence, data mining, computer vison, image processing and robotics.
\end{IEEEbiography}
\begin{IEEEbiography}
[{\includegraphics[width=1in,height=1.15in,clip,keepaspectratio]{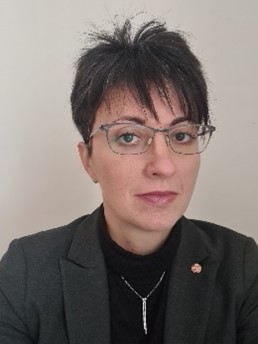}}]{Laura Moretti}
PhD in Infrastructures and Transportation, is Associate professor at Sapienza University of Rome - Department of Civil, Building and Environmental Engineering. Her research interests are risk analysis, life cycle assessment, management and construction of transport infrastructures. Her teaching activities are about Safety of road works, Design of transportation during emergencies, Road infrastructures, and Concrete pavements for bachelor, master, and second-level master degrees. She is involved in several research on life cycle assessment, green public procurement, airport safety, environmental impact, design, construction and maintenance of roads and airports. She is the author of more than 70 articles published in conference proceedings and international journals and she is a member of scientific committees at several conferences.
\end{IEEEbiography}
\begin{IEEEbiography}
[{\includegraphics[width=1in,height=1.15in,clip,keepaspectratio]{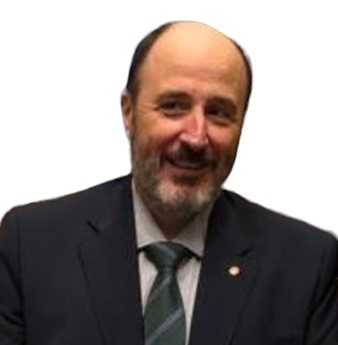}}]{Giuseppe Loprencipe}
received the M.S. degree in Civil engineering from the Sapienza University of Rome - Italy, in 1994, and the Ph.D. degree in Transportation Infrastructures from the Sapienza University of Rome - Italy, in 2000. He is currently a full Professor at the Sapienza University of Rome. His research interests include pavement management, road, railway, and airport engineering. He is a member of the SIIV (Italian Society of Road Infrastructures) governing council and a member of the commission of the Italian Ministry of Infrastructure and transportation on climate change.
\end{IEEEbiography}
\begin{IEEEbiography}
[{\includegraphics[width=1in,height=1.15in,clip,keepaspectratio]{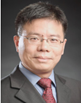}}]{Weihua Sheng (Senior Member, IEEE)}
received the B.S. and M.S. degrees in electrical engineering from Zhejiang University, China, in 1994 and 1997, respectively, and the Ph.D. degree in electrical and computer engineering from Michigan State University, East Lansing, MI, USA, in 2002. He is currently an Associate Professor with the School of Electrical and Computer Engineering, Oklahoma State University, Stillwater, USA. His current research interests include wearable computing, and intelligent transportation systems.
\end{IEEEbiography}
\begin{IEEEbiography}
[{\includegraphics[width=1in,height=1.15in,clip,keepaspectratio]{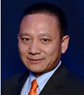}}]{Kelvin C. P. Wang}
received the B.S. degree from Southwest Jiaotong University, China, the M.S. degree from Beijing Jiaotong University, China, and the Ph.D. degree from Arizona State University. His professional career started at Arizona DOT in 1989, where he has been a university faculty since 1993. He is currently a full Professor of civil engineering with Oklahoma State University Stillwater, where he holds the Dawson Chair. He received the ASCE 2011 Frank M. Masters Transportation Engineering Award and the ASCE 2018 Turner Award. He was
the President of the Transportation and Development Institute of the American Society of Civil Engineers for FY 2017.
\end{IEEEbiography}

\end{document}